\documentclass{article}
\usepackage{aaai}
\usepackage[square,comma]{natbib}
\usepackage{mathptmx}
\usepackage{url}
\usepackage{xcolor}
\usepackage{graphicx}  %Required
\newtheorem{exampl}{Example}
\usepackage{amsmath}

\newcommand{\tuple}[1]{\langle #1 \rangle}

\newcommand{\LPimplies}{{\operatorname{~:-~}}}

\title{Comparing Aggregators for Relational Probabilistic Models}

\author{Seyed Mehran Kazemi, Bahare Fatemi, Alexandra Kim, Zilun Peng, \\[0.3em]\Large\bf Moumita Roy Tora, Xing Zeng, Matthew Dirks \and David Poole\\[0.3em]
Dept. of Computer Science, University of British Columbia
} % LEAVE BLANK FOR ORIGINAL SUBMISSION.
\begin{document}

\maketitle

\begin{abstract}
Relational probabilistic models have the challenge of aggregation,
where one variable depends on a population of other
variables. Consider the problem of predicting gender from movie ratings; this is
challenging because the number of
movies per user and users per movie can vary greatly. Surprisingly, aggregation is not well understood. In this paper, we show that existing relational models (implicitly or explicitly) either use simple numerical aggregators that lose great amounts of information, or correspond to \emph{naive Bayes}, \emph{logistic regression}, or \emph{noisy-OR} that suffer from overconfidence. We propose new simple aggregators and simple modifications of existing models that empirically outperform the existing ones. The intuition we provide on different (existing or new) models and their shortcomings plus our empirical findings promise to form the foundation for future representations.
\end{abstract}

\section{Introduction}
Much research in machine learning and reasoning under
uncertainty is application-driven; the aim is to predict a distribution as well as possible given all of the
information available. In science, it is normal to have idealized
experiments where research concentrates on one aspect, and extraneous
variables (confounders) are eliminated as much as possible. This paper
examines the problem of aggregation in relational probabilistic
models, which we try to study without confounders such as other
properties or relations.

Probabilistic reasoning and learning about objects and relations has been studied
under the umbrella of statistical relational learning \citep{Getoor:2007aa}
or statistical relational AI \citep{DeRaedt2016}. 
The models, which we refer to generically as relational probabilistic models,
are characterized by weights over first-order formulae (where not all
variables are quantified), 
and typically mean the grounding,
where unquantified (free) variables are substituted by the constants in the language in all
possible ways. The quantification allows for parameter sharing or
weight tying, where the different instances share the same
weights/parameters.

These models can be directed (such as in various instances of
probabilistic logic programming
\citep{reason:Poole93g,Sato:1997aa,De-Raedt:2007aa}) or undirected
(such as in Markov logic networks \citep{Domingos:2009}).

Relational models are typically defined in terms of the grounding of the model, where there is a random variable instance for each assignment of an object to each logical variable. The neighbors are those instances that are in a (weighted) formula.
This could be seen as making relational models just like graphical models (with some parameter sharing / weight tying).
One way in which relational models differ from standard graphical models is when a variable can have an unbounded number of neighbors in the graphical model, and the number of neighbors can vary from instance to instance. This problem of aggregation is a major subproblem in relational models and occurs when a variable, property or relation is in a formula with a property or relation that has extra logical variables. This paper investigates the simplest non-trivial form of such aggregation, where a property depends on a relation with one extra logical variable, and other confounders are deliberately ignored.

More than a decade ago, \citet{perlich2003aggregation} made a first step toward better understanding of the aggregation problem for relational models. They concluded that the existing relational models to that date were relying on poor aggregators, and that a better understanding of why each model performs well or badly on different domains is required. Since then, several relational models have been developed \citep{neville2003learning,De-Raedt:2007aa,natarajan2008learning,Domingos:2009,natarajan11mlj,kimmig2012short,Kazemi:2014aa}. 
The types of aggregators used in these models as well as the learning issues arising when aggregating, however, are still poorly understood.

In this paper, we investigate the types of aggregation used in current relational models. All these models (explicitly or implicitly) rely on either numerical aggregators such as \emph{mean}, \emph{count}, \emph{proportion}, etc. or correspond to a \emph{naive Bayes}, \emph{logistic regression}, or \emph{noisy-OR} model. We discuss the problems with relying on any of these and show as a consequence that none of the existing relational models work entirely satisfactorily for aggregation. We then
propose new (simple) aggregators that work better and have the potential to form the central component of future relational representations. 

Note that this is an experimental paper; we deliberately create subproblems that only include aggregation. This is not an application paper; we are not trying to do as well as we can in those domains, which would entail using all information, but would then interfere with testing methods for aggregation. Aggregation is only one of the tools needed for applications, and it is important to understand its characteristics in isolation as well as combined with other methods. This is also an exploratory paper in that we try many qualitatively different methods to see which ones work; we are not trying to find the definitively best answer, which can only be done when the space of methods is better defined.

\section{A Motivating/Running Example}
As a motivating/running example, consider the problem of predicting gender
from movie ratings.
The MovieLens movie rating dataset(s)
\citep{Harper:2015} have relations on $\tuple{u,i,r,t}$ for
user $u$, item $i$ (where, here, items are movies), rating $r$ and
timestamp $t$, which we will write as relation $rating(U,I,R)$,
ignoring the timestamp during training and testing. In some of the datasets,
the gender of each user is recorded, which we write as relation
$gender(U)$. 
In popular culture there is a hypothesis that ratings depend on gender (there are some movies that appeal to females and some to males); this hypothesis is also supported by \cite{weinsberg:2012}. It aligns with our results as well, but many of the existing methods cannot extract that signal. In
a probabilistic model, because users rate multiple movies, there needs
to be some aggregation of the movies a user has rated to predict the
gender of the user. 

In particular, we initially use the first 60,000 ratings of the ml-100k dataset, 
which we call MovieLens 60K. 
We use the gender of the users who rated the
first 40,000 ratings as the training set,
and predict the gender of the users of the remaining 20,000 ratings\footnote{See
  \url{https://grouplens.org/datasets/movielens/} The ratings used are
those with a timestamp of  884673930 or less, and the gender is those
that rated on or before timestamp  880845177.}. This subset contains 1451 movies, 419 users used for training
and 171 test users.
This is more
challenging (and more interesting) than the full MovieLens dataset because it
includes users with few ratings (whereas the full dataset has these
removed). We use this as an ongoing example to understand and debug various aggregation schemes.

The challenge is (mainly) because the number of ratings per person is
extremely varied. 
In this dataset the number of ratings per user ranges from 1
to 559, the mode is at 20 (with 18 users giving a rating of 20), the
mean number of ratings per user is 102. Two medians are of interest: half of
the users rated 66 or fewer movies, but half of the ratings were 
by users who rated 163 or fewer movies.   
Similarly, the number of ratings per movie ranges from 1 to 375, with 126 movies with just 1 rating. 
There are 297 movies that no females rated, and 28 movies that no males rated.

To keep things simple, we only consider whether a rating was greater than or equal to 4, which we write as $r_{\geq 4}$, or the rating was less than 4, which we write as $r_{<4}$.
There is a subtle issue about what to do with ratings that do not appear in the database. 
Some of the methods treat these as a third value, and some as being unobserved (although many of them can be coerced into one way or the other). We will be explicit about how unobserved ratings are treated.

There are two main challenges that arise:
\begin{itemize}
\item Regularization: with many low counts, we need 
to regularize. This needs to be taken into account in the inference as
well as in the learning, and so needs to be explicitly part of the representations.
\item Independence: movies do not provide independent evidence. For example, one might imagine that ratings for \emph{Godfather 1} and \emph{Godfather 2} are strongly correlated.  With many movies rated by some people, determining dependency is important. With low counts (by each user and for each movies), determining dependence is difficult.
\end{itemize}
The challenges do not disappear with more data; larger datasets have more movies and more users. It is not that users watch more movies, although larger datasets do capture a higher proportion of the viewers of popular movies.

\section{Background and Notation}
Relational models are not just big non-relational models because of
parameter tying and because variables can depend on varying numbers of
other variables. In this section, we define some basic terminologies as well as a subset of first-order logic and logic programs that are the basis for relational models, and briefly describe the relational models used in the rest of the paper.

\subsection{Finite-domain, Function-free, First-order logic}
A \textbf{population} is a finite set of \emph{objects}. The size of a population corresponds to the number of objects in its set. 
A \emph{logical variable (logvar)} is typed with a population and is written in upper-case. 
For a logvar $X$, we show the size of the population associated with it as $|X|$.
\textbf{Constants} denoting objects are written in lower-case. A \textbf{term} is a logvar or a constant.
An \textbf{atom} is of the form $f(t_1, \dots, t_k)$ where $f$ is a predicate or a functor, and each $t_i$ is a term. 
An atom is \textbf{ground} if all its terms ($t_i$) are constants. A \textbf{literal} is an atom or its negation. 
A \textbf{formula} $\varphi$ is a literal, the negation $\neg \varphi$ of a formula, a disjunction $\varphi_1 \vee \varphi_2$ of formulae, or a conjunction $\varphi_1 \wedge \varphi_2$ of formulae. A \emph{theory} is a set of of formulae that implicitly form a conjunction.
A \textbf{world} is a value assignment to all ground atoms in a theory.

\subsection{Logic programs}
A \textbf{rule} is of the form $h(\dots) \LPimplies \varphi$ where $h(\dots)$ is an atom representing the \textbf{head} and $\varphi$ is a formula representing the $body$ of the rule. 
A \textbf{fact} is a rule where the body $\varphi$ is true. 
A \emph{logic program (LP)} is a set of rules. 
Every variable in a LP that is not implied to be true is assumed to be false. This assumption is known as \emph{closed-world assumption (CWA)}. 
Every logical variable appearing in $\varphi$ but not in $h(\dots)$ is assumed to be existentially quantified. 
For instance having a rule $s \LPimplies r(X)$ is interpreted as $s \LPimplies (\exists X: r(X))$, meaning $s$ is true if $r$ is true for at least one object.

\subsection{Markov logic networks}
Markov logic networks (MLNs) \citep{Domingos:2009} are one of the most well-known and widely-used relational models. MLNs use \emph{weighted formulae} to define joint distributions over ground atoms.

A \emph{weighted formula (WF)} is a pair $\left\langle w, F \right\rangle$ where $F$ is a formula and $w$ is a weight. An MLN is a set of WFs, where the probability of any world
is proportional to the exponent of the sum of the weights of the groundings of the formulae that are true in the world.

\subsection{Relational logistic regression}
Relational logistic regression (RLR) \citep{Kazemi:2014aa} is the directed analogue of MLNs. RLR defines a conditional probability distribution over a child atom using WFs consisting of atoms from the child's parents. The formulae of WFs can be viewed as features whose values are computed by counting the number of times the formula is $True$ based on the parent values. The probability of child being $True$ is the $Sigmoid$ of the weighted sum of these features. \citet{fatemi2016learning} show how hidden object features with continuous values can be also included in WFs and learned directly for each object during training.
\citet{Kazemi:2014aa} show that not only several well-known explicit aggregators (such as \emph{noisy-OR}, \emph{noisy-AND}, $average>t$, $mode=t$), but any other aggregator that is a polynomial of counts can be modelled or approximated using RLR.

\subsection{Problog}
Problog \citep{De-Raedt:2007aa} is a probabilistic logic program where a probability is assigned to each fact. These probabilities define a probability distribution over (non-probabilistic) logic programs. The probability of a ground atom being $True$ in a Problog program is the sum of the probabilities of the logic programs that entail the ground atom.

\subsection{RDN-Boost}
\citet{natarajan11mlj} learn multiple relational probability trees using gradient boosting and use all learned trees to make predictions about unseen cases. Their boosting method is known as RDN-Boost.

Relational probability trees \citep{neville2003learning} can be viewed as standard decision trees where the decision nodes may include thresholds on aggregation functions such as \emph{count}, \emph{average} and \emph{proportion}. In our running example, for instance, a node may split the users based on whether they have rated more than $10$ movies or not.

\section{The Aggregators}
In this section we describe the aggregators and show how they work for the running example. The atoms $g(U)$ is the gender of user $U$,   $r_{\geq 4}(U,M)$ means the rating of user $U$ for movie $M$ is greater than or equal to $4$, and and $r_{< 4}(U,M)$ means rating less than $4$.  

\subsection{Existing Explicit Aggregators} \label{explicit}
Existing explicit aggregators in the literature (see e.g., \citet{reason:HorPoo90a,Friedman:1999lr,neville2003learning,Neville:2007aa,Kisynski:2009ab,Kazemi:2014aa}) include \emph{OR}, \emph{AND}, \emph{noisy-OR}, \emph{noisy-AND}, \emph{logistic regression} and numerical functions such as \emph{average}, \emph{count}, \emph{median}, \emph{mode}, \emph{max/min} and \emph{proportion}.

\emph{OR} and \emph{AND} (corresponding respectively to whether the user has rated at least one movie and whether a user has rated all movies in our running example) will obviously perform badly. \emph{Noisy-OR} and \emph{noisy-AND} are both monotonic functions: with noisy-OR, more observations can only increase the probability of a certain class, and with noisy-AND, more observations can only decrease the probability of a certain class. These two aggregators may be expected to perform badly on our running example as each observation (each movie rated by a user) may only increase (or only decrease) the probability of a class.

\emph{Logistic regression} solves the monotonicity problem of \emph{noisy-OR} and \emph{noisy-AND} functions as each observation may increase or decrease the probability of a certain class. For instance, if $I_1$ has been mostly rated by males and $I_2$ has been mostly rated by females, then observing a new user has rated $I_1$ may increase
the probability that this user is male
and observing they have rated $I_2$ may decrease 
the probability that this user is male. 
However, as pointed out by \cite{poole2014population}, as the number of ratings increases, a logistic regression model tends to become over-confident about its predictions (predicting with probabilities near zero or one) which results in poor performance in terms of log loss.

Numerical functions such as \emph{average}, \emph{count}, \emph{median}, \emph{mode}, \emph{max/min} and \emph{proportion} may provide small hints. In our running example, for instance, on average males may have rated more movies than females, thus thresholding \emph{count} may provide a small hint for prediction. However, all of these functions lose great amounts of information as they do not consider which movies females or males like better. 

\subsection{Existing Implicit Aggregators} 
Here we consider some well-known relational learning models and how they do aggregation.
\subsubsection{MLNs and RLR}\label{MLN-secn}
An MLN model for our running example is as follows:
\begin{align}
&\langle w_0, ~g(U)\rangle\nonumber\\
&\langle w_1, ~ r_{\geq 4}(U,M) \land g(U)\rangle \label{MLN-eqn}\\
&\langle w_2, ~ r_{<4}(U,M) \land g(U)\rangle\nonumber
\end{align}
We assume the closed world assumption (having no rules for the negations of $r_{\geq 4}$ and $ r_{<4}$ means that unobserved ratings have no effect on the probability of the models). With this assumption, the MLN model is equivalent to an RLR model with the same weighted formulae.
The value of $w_0$ reflects the prediction for no observed ratings, 
weight $w_1$ the number of ratings that are 4 or greater
and $w_2$ the number of ratings that are less than 4.
Note that replacing $\land$ (and) with $\lor$ (or) or $\rightarrow$ (implies),
or adding rules containing $\neg g(U)$,
does not change the model or what can be represented (but the weights change). 

It can be shown that from this model
\begin{align}
P&(g(U)\mid rs) \nonumber\\
&= sigmoid(w_0 + w_1*npr(U) + w_2*nnr(U))  \label{mln-eqn}
\end{align}
where $npr(U)$ is the number of positive ($\geq 4$) ratings for user $U$ in rating set $rs$, and $nnr$ is the number of negative ratings ($<4$) in rating set $rs$. 

We would not expect this model to work well. As shown by \cite{poole2014population}, as the number of ratings for a user increases ($npr(U)$ and $nnr(U)$ increase), the probability of $g(U)$ will approach 0 or 1, unless $w_1$ and $w_2$ are approximately zero (or happen to cancel out). Thus we would expect that to avoid extreme predictions, $w_1$ and $w_2$ go to zero, which means that the model effectively ignores the ratings in the prediction of gender.

We also tested models with a hidden variable:
\begin{align*}
&\langle w_0, ~g(U)\rangle\\
&\langle w_1, ~ r_{\geq 4}(U,M) \land h(U)\rangle\\
&\langle w_2, ~ r_{<4}(U,M) \land h(U)\rangle\\
&\langle w_3, ~g(U) \land h(U)\rangle\\
&\langle w_4, ~g(U) \land \neg h(U)\rangle
\end{align*}
where, again, adding extra formulae involving $h$, $r$ and $g$ does not affect the model. In this case, the $RLR$ model and the $MLN$ model are different (in how $h$ is marginalized). We expected the hidden variable to act as a regularizer, which makes the model make none-extreme predictions for $g(U)$ even if the $h(U)$ values become extreme. This model, however, did not perform better than the one with no hidden variables.

Another model we tested is to have different weights for each movie as in the following schema:
\begin{align}
&\langle w_0, ~g(U)\rangle\nonumber\\
&\langle w_{i1}, ~ r_{\geq 4}(U,M_i) \land g(U)\rangle \label{MLN-schema-eqn}\\
&\langle w_{i2}, ~ r_{<4}(U,M_i) \land g(U)\rangle\nonumber
\end{align}
for each movie $M_i$. 
Assuming all ratings are observed (the closed-world assumption), this is equivalent to an RLR model for $g(U)$ with the following WFs:
\begin{align}
&\langle w_0, True\rangle\nonumber\\
&\langle 1, r_{\geq 4}(U,M) * h_1(M)\rangle \label{RLR-schema-eqn}\\
&\langle 1, r_{< 4}(U,M) * h_2(M)\rangle\nonumber
\end{align}
where $h_1(M)$ and $h_2(M)$ are continuous hidden atoms whose value for each movie can be directly learned during training.
We report the results for this model in our experiments.

\subsubsection{Problog}
For predicting the gender of users given their ratings in our running example, a Problog program may be as follows:
\begin{align*}
&w_0::g(U). \\
&w_1::g(U)~ :- ~r_{\geq 4}(U,M).\\
&w_2::g(U)~ :- ~r_{<4}(U,M).
\end{align*}
which is equivalent to the following program where weights are only assigned to facts:
\begin{align*}
&w_0::n_0(U).\\
&w_1::n_1(U,M).\\
&w_2::n_2(U,M).\\
&g(U) ~:- ~ n_0(U).\\
&g(U) ~:- ~ r_{\geq 4}(U,M)\wedge n_1(U,M).\\
&g(U) ~:- ~ r_{<4}(U,M)\wedge n_2(U,M).
\end{align*}
where $n_0(U)$ and $n(U,M)$ are noise variables. Again, this model corresponds to taking into account only the number of movies rated by a user (i.e. the explicit aggregator \emph{count}), and building a model using just this feature with weight $w$.

In this case,
\begin{align}
P&(g(U)\mid rs) \nonumber\\
&= 1-(1-w_0)*(1-w_1)^{npr(U)}*(1-w_2)^{nnr(U)}  \label{problog-eqn}
\end{align}

\subsubsection{RDN-Boost}
RDN-Boost \citep{natarajan11mlj} learns regression trees whose intermediate nodes may contain thresholds over explicit functions mentioned in subsection~\ref{explicit}. Therefore, the type of aggregation in RDN-boost is a combination of thresholding functions. 
We use RDN-Boost as a test for the standard explicit exaggerators. If the standard aggregators are useful, then RDN-Boost should perform well. Our test results show that RDN-Boost does not perform well, which we take as evidence that the underlying aggregations are not extracting useful information (as one might expect).

\subsection{New Aggregators}
In this section, we propose some new aggregators.  We tried and rejected many aggregators; the ones presented are the ones that are best for each type of aggegator.

\subsubsection{Treating each movie as a dataset}
One intuition is that the movies can be seen as datasets in themselves. A movie with $n$ ratings can be seen as $n$ data points about the users who rated this movie. The idea behind this approach is similar to rule combining of \citet{natarajan2008learning}, where each parent independently predicts the target and then these independent predictions are combined. However, we do not tie the parameters for parents, and we do not consider intermediate hidden variables for each parent's prediction that should be learned using EM.
One way this can be used is to use all of the people who rated the movies that $u$ rated:
\begin{align}
P_1&(female(u)\mid E) \nonumber\\
&= \frac{\displaystyle c+\sum_{\{i:rated(u,i)\}} \# u': rated(u',i)\wedge female(u')}{\displaystyle 2c+\sum_{\{i:rated(u,i)\}} \# u': rated(u',i) \label{p1defn}}
\end{align}
Where the denominator only includes ratings by users whose gender is known, and $c$ is a pseudo-count (beta/Dirichlet prior), which is set in our experiments by 5-fold cross validation (and was often quite large). 

Counting only the users that rated the same as $u$ worked slightly worse (presumably because there were fewer datapoints for each user). 
Using more informed prior counts (which predict the training average for users that rated no movies) also worked worse. We will not consider these variants here.

The second is to average over all movies that $u$ rated:
\begin{align}
P_2&(female(u)\mid E) \nonumber\\
&= \frac{1}{\#i:rated(u,i)}\sum_{\{i:rated(u,i)\}} avefem(i) \label{p2defn}
\end{align}
where
\begin{align*}
avefem(i) =\frac{c+\# u': rated(u',i)\wedge female(u')}{2c+\# u': rated(u',i)}
\end{align*}
The value of $c$ was set by 5-fold cross validation, but was often around 1 (which corresponds to Laplace smoothing).

\subsubsection{Limiting the number of neighbors}
One ``obvious'' solution is a generative model where the gender produces the ratings, which results in a Naive Bayes model under the assumption that ratings are independent given the gender.
Naive Bayes does not work well because the independence assumption is not appropriate;
the ratings are not independent given the gender. 

Naive Bayes is optimal if there is a single rating (assuming others are missing at random). 
However, the model becomes extremely overconfident (probabilities approach 0 or 1) as the number of ratings increases. 
The independence, however, is approximately correct; it works when there are a few ratings, just not hundreds. 
One way to fix this is to limit the number of movies considered for each user; instead of using all of the ratings, a subset can be used. 
The movies with very few ratings do not provide a useful signal (as they need to be regularized too much), and the movies with very many ratings also turn out to be not very useful, because they are just popular independently of gender. 
We tested selecting $k$ ratings at random for each user. 
The value of $k$ was selected by 5-fold cross validation.  
The problem with this method is that the answer is sensitive to which $k$ movies are selected, and so the prediction is the average of many selections, which makes predicting slow or noisy.

\subsubsection{MLN/RLR with relational dropout}
Dropout \citep{srivastava2014dropout} is a regularization technique designed mainly to avoid overfitting in deep neural networks. 
In training a deep network with dropout, 
each iteration will randomly select a proportion (e.g., 30\%) of neurons and turn them off, thus forcing the network to make predictions using only the other neurons. 
For making a prediction about an unseen case, the network is used several times (each time a proportion of neurons are selected randomly and turned off) and the average is reported. This allows the network to average over multiple configurations.

We utilize the idea of dropout to avoid making over-confident predictions in MLN/RLR. 
During training the MLN/RLR model in Eq~(\ref{MLN-schema-eqn})~and~(\ref{RLR-schema-eqn}), in each iteration we turn off a part of the observations randomly and make predictions using only the rest of the observations. 
However, instead of keeping a \emph{proportion} of the observations on (like standard dropout), we keep a fixed number (e.g., only considering 10 random movies rated by a user) of them on. The reason why we keep a fixed number instead of a proportion of the observations is that the number of observations (and thus a proportion of them) varies per object. In our running example, for instance, each user has rated a different number of movies. In test time, we report the average over multiple predictions of the model, where each time a fixed number of observations are selected randomly and kept on. 

Note that similar to naive Bayes with limited neighbors, averaging over multiple fixed-length subsets of the movies makes the predicting slow or noisy. 

\begin{table*}
\begin{center}
\begin{tabular}{c|c|c|c|c|c|c}
&\multicolumn{3}{c|}{Target object}&\multicolumn{2}{c|}{Other object}&\multicolumn{1}{c}{Ratings}\\
Dataset & Type & \#Train  & \#Test & Type & Count & Count  \\\hline
MovieLens-60K & User & 419 & 171 & Movie & 1511 & 60,000 \\\hline
MovieLens-600K & User & 2665 & 1416 & Movie & 3575 & 600,000 \\\hline
Yelp & Business & 3525 & 882 & User & 2108 & 38,327
\end{tabular}
\end{center}
\caption{Summary of the datasets used for experiments.}
\label{datasets-summary}
\end{table*}

\subsubsection{Matrix Factorization}
Matrix factorization techniques have proved effective especially in recommendation systems \citep{Koren:2009aa}. 
The idea is to factorize a relation matrix over two types of objects (e.g., users and movies) into latent features for each type of objects such that the relation matrix can be approximated using the latent features. \citet{Koren:2009aa} use matrix factorization for movie recommendation and factorize the rating matrix as follows:
\[\hat{R}_{ui} = \mu + b_u + b_i + \sum_f P_{uf}*Q_{fi} \]
where $\hat{R}_{ui}$ is the predicted rating of user $u$ for movie $i$, $\mu$ is a bias, $b_u$ and $b_i$ are user and movie biases, and $P_{uf}$ and $Q_{fi}$ are the f-th latent property of the user $u$ and movie $i$ respectively.

When the ratings are in binary (e.g., greater than equal to 4 or less than 4), the model can be trained to predict:
\[P(\hat{D}_{ui}) = sigmoid(\mu + b_u + b_i + \sum_f P_{uf}*Q_{fi}) \]
where $D_{iu}$ is true when the rating for user $u$ on item $i$ is 4 or greater and is false when it is less than 4. The same training algorithm as \citet{Koren:2009aa} can also be used here to minimize log loss.

While matrix factorization method might not be new, it has not be used explicitly for aggregation.

One way to use matrix factorization for aggregation is to factorize the relation matrix into latent features, then use the latent features of the desired objects as  features and build a model over these features to make predictions about them. 
For our running example, we train a logistic regression on
$b_u$ and $P_{uf}$ for each $f$ to predict the gender of the users, which takes the following form:
\begin{align}
P&(female(u)\mid b_u, P_u) \nonumber\\
&= sigmoid(w + w'* b_u + \sum_f w_f * P_{uf}) \nonumber
\end{align}
where $P_{uf}$ is, as previously mentioned, the f-th latent property of the user $u$. This was th method reported as ``matrix factorization'' in the results.

Note that we also tried a number of other methods such as forcing one of the features to be gender, but none of them worked as well as what is described here. We had also tried using Bayesian probabilistic matrix factorization, a graphical model variant of matrix factorization proposed by \cite{salakhutdinov2008bayesian} which provides automatic complexity control. But it mostly had a similar performance as the regular matrix factorization method so we did not include it in our final comparison. 

One thing that works well is to initialize one of the latent features to correlate with gender (e.g., to initialize the feature with $+2$ for females and $-2$ for males, and a random value in (-0.1,0.1) for other objects and for other features).

Instead of learning the latent features over the rating matrices, and using them to predict gender, we also tried learning latent features that predict both ratings and gender simultaneously. 
This, however, did not do well. We identified two reasons for the poor performance of this model:
\begin{itemize}
\item Since there are many more ratings than genders, the latent features tend to be more inclined toward predicting ratings.
\item One of the latent features of the users may become identical to the gender of the user (thus completely over-fitting to train data), while the rest of the latent features predict the ratings.
\end{itemize}
Learning latent features that jointly predict multiple targets has been used extensively for knowledge base completion (see e.g.,  \citep{nickel2012factorizing,bordes2013translating,nickel2016holographic,trouillon2016complex,nguyen2017overview}). However, to the best of our knowledge only \cite{nickel2012factorizing} considers the properties of objects; the other ones only consider relationships and ignore object properties.
We believe knowledge base completion tools should take the two issues we identified into account if they want to include object properties along with object relations.

\section{Empirical Evaluation}\label{results}
\subsection{Evaluation Measures}
We compare the algorithms on the test set using two measures that are suited for probabilistic predictions for Boolean variables.
\begin{itemize}
\item mean squared error (MSE):
\[\frac{1}{|T|}\sum_{e \in T} (\widehat{p}(e)-a(e))^2\]
where $T$ is the set of test examples, $\widehat{p}(e)$ is the prediction on example $e$ and $a(e)$ is the actual value for example $e$. Predicting $\widehat{p}(e)=0.5$ always has an error of 0.25.
\item log loss (LL), the negative of log likelihood, divided by the number of examples:
\[-\frac{1}{|T|}\sum_{e \in T}a(e)*\log_2 \widehat{p}(e) + (1-a(e))*\log_2 (1-\widehat{p}(e))\]
which is undefined if either of the logarithms is given a number less
than or equal to 0. We use base-2 logarithms so the answer can be interpreted in bits. Predicting 0.5 has a log loss of 1.
\end{itemize}
Log loss puts a larger penalty on extreme predictions that are incorrect, which is appropriate if using the predictions for making decisions.

\subsection{Datasets}
The datasets used in our experiments are:
\begin{enumerate}
\item Our running example: MovieLens 60K.
\item A subset of the larger MovieLens 1M dataset \citep{Harper:2015} that we call MovieLens 600K. The 1M dataset is the largest MovieLens dataset that contains gender. 
As in the 60K data, we used only the first 60\% of the ratings, 
of which we used the users who had rated the movies in the first 40\% 
 as the training set and the remaining users as the test set. This is more challenging than using the whole dataset as it is less cleaned (the complete dataset removes users with few rating, but we include these), and because predicting future users is an extrapolation task that should be more challenging than interpolation.
\item A business prediction task extracted from the Yelp challenge dataset\footnote{\url{https://www.yelp.com/dataset_challenge}} for predicting the type of food offered in a restaurant. We considered all restaurants offering either Chinese or Mexican food (but not both), and restricted the users to those who have rated at least 10 of these restaurants. The number of ratings per restaurant ranges from $0$ to $216$.
\end{enumerate}
Table~\ref{datasets-summary} represents a summary of the three datasets.

\subsection{Results}

For many of these, we tried to use the standard software, but often this did not work for our datasets. For the software we wrote, unless otherwise specified, we typically used gradient descent, trained enough to reach a local minimum on the test set. Hyper parameters (typically regularization parameters) were trained by 5-fold cross validation.

The results on the three datasets are shown in Table~\ref{results-fig}.

\begin{table*}
\begin{center}
\begin{tabular}{l|c|c|c|c|c|c|}
&\multicolumn{2}{c|}{MovieLens 60K}&\multicolumn{2}{c|}{MovieLens 600K}&\multicolumn{2}{c|}{Yelp}\\
Method & MSE & LL  & MSE & LL & MSE & LL  \\\hline
Predict 0.5 & 0.250 & 1.000 & 0.250 & 1.000& 0.250 & 1.000\\
Training average & 0.216 & 0.900 & 0.204 & 0.864 & 0.236 &  0.960 \\
MLN/RLR (no hidden) (Eq (\ref{MLN-eqn})) & 0.211 & 0.881 & 0.204 & 0.863 & 0.234 & 0.953 \\
Alchemy (with model of Eq (\ref{MLN-schema-eqn}))  &	0.220 & 	1.149 & 	0.159 &	0.853 & 	\textbf{0.198} & 	0.871\\
Problog (noisy-or) & 0.211 & 0.882 & 0.203 & 0.861 & 0.233 & 0.952 \\
RDN-Boost  & 0.215 & 0.899 & 0.204 & 0.864 & 0.234 & 0.953 \\
Movies as a dataset (Eq (\ref{p1defn})) & 0.207 & 0.868 & 0.195 & 0.831 & \textbf{0.198} & \textbf{0.833} \\
Average of each movie as dataset (Eq (\ref{p2defn})) & 0.209 & 0.875 & 0.190 & 0.811 & \textbf{0.197} & \textbf{0.831} \\
MLN/RLR with relational dropout & \textbf{0.199} & 0.849 & \textbf{0.148} & \textbf{0.668} & \textbf{0.198} & 0.837 \\
Matrix Factorization & \textbf{0.200} & \textbf{0.844} & 0.193 & 0.824 & 0.236 & 0.959 \\
\end{tabular}
\end{center}
\caption{Results of various aggregators (lower is better)}
\label{results-fig}
\end{table*}

For the MLN and RLR results, we tried Alchemy 2.0 \citep{kok2009alchemy}.  To keep the evaluations consistent with the other methods, we directly optimized Equation (\ref{mln-eqn}) by stochastic gradient descent. Our results are consistent with Alchemy, but were much faster to obtain (but admittedly our programs were not as general as Alchemy).

The weights found for the model (\ref{MLN-eqn}) were approximately $w_0=0.999$, $w_1=0.0023$ and $w_2=-0.0043$, which indicates that ratings 4 and over were positive evidence for being a female and, ratings of less than 4 were negative evidence for being female. The parameters $w_1$ and $w_2$ are very small as they are multiplied by the number of ratings. The model with a hidden variable did not perform better (but found many different parameterizations all with essentially the same error).

Problog ran for a few days to train the parameters and then crashed (even for the 60k dataset). To judge the results it would have given, we directly optimized Equation (\ref{problog-eqn}) by stochastic gradient descent, which works well for this particular problem but may not be a general solution. The program found negative probability for $w_2$, for the same reason that the MLN found a negative weight for its corresponding weight. This could indicate that this model is not appropriate, but note that negative probabilities have been advocated as being needed for probabilistic logic programs \citep{Buchman:2016aa}.

RDN-Boost did not perform very well for this task. This is because the aggregation primitives used in their code could not extract useful features. We did not try using RDN-Boost with other aggregation methods as their base learners. It would be interesting to try RDN-Boost with the other aggregators that we have found to be useful.

The next model is the MLN/RLR model using a bias and two weights per item, following the clauses of formula (\ref{MLN-schema-eqn}). The table shows the results for Alchemy (using an equivalent formalism using disjunctions, as Alchemy seems to work better with disjunctive formulae). The log likelihood is terrible for the 60K dataset because it becomes overconfident in predictions that are wrong, which are penalized much more in log likelihood than in squared error.

The use of the movies as datasets improves over the previous methods. These results just considered whether the movie was rated, and not what the rating was. Using the actual rating (or whether it was $\geq 4$ or $<4$) performed slightly worse.

Limiting the number of neighbors in relational dropout makes these competitive, which gives evidence that the problem is overconfidence. The overconfidence problem, and the ability of relational dropout to overcome it to a certain amount, can be  especially viewed for ML-600K where relational dropout outperforms all other methods. Limiting the number of parents is a simple idea to solve the problem, and so we think that there should be ways to improve such methods.

\section{Conclusion}
One aspect of relation models that differs from standard graphical models is the need for aggregation where the number of neighbors varies for the grounding of the atoms for different variables. The difference can be orders of magnitude. We showed empirically that the current models cannot handle this diversity, and using  simple methods (or by simple modifications of existing methods) we can do better than the existing relational models. We discussed the reasons why the current models perform poorly.

In this paper we have made an attempt at understanding a major  subproblem that arises in statistical relational learning. We hope that building on the foundations of good solutions to subproblems will form the foundations for the next generation of representations.

\bibliographystyle{dpbib3}
\bibliography{string,reason}
\end{document}